\documentclass[lettersize,journal]{IEEEtran}
\usepackage[linesnumbered, ruled]{algorithm2e}
\usepackage{amsthm,amsmath,amssymb}
\usepackage{array}
\usepackage[caption=false,font=normalsize,labelfont=sf,textfont=sf]{subfig}
\usepackage{textcomp}
\usepackage{stfloats}
\usepackage{url}
\usepackage{verbatim}
\usepackage{graphicx}
\usepackage{booktabs,color}
\usepackage{cite}
\usepackage[affil-it]{authblk}
\usepackage{stackengine} 
\usepackage{mathtools}
\usepackage{mathrsfs}

\usepackage{indentfirst} 
\usepackage{multirow}
\usepackage[backref=page]{hyperref}
\usepackage{cleveref}
\usepackage{xcolor,eucal}
\usepackage{makecell}

\usepackage{pifont}
\usepackage{epsfig}

\let\oldnl\nl% Store \nl in \oldnl
\newcommand{\OUR}{AdaBLDM}
\newcommand{\nonl}{\renewcommand{\nl}{\let\nl\oldnl}}% Remove line number for one line
\graphicspath{{figs/}}

\newcommand{\hanxiR}{\mathbb{R}}
\newcommand{\II}{\text{I}}

\newcommand{\bb}[1]{\mathbf{#1}}

\newcommand{\NL}{\\}

\SetKwRepeat{Do}{do}{while}%
\begin{document}

\title{A Novel Approach to Industrial Defect Generation through Blended Latent Diffusion Model with Online Adaptation}

\author[1,2,$\dagger$]{Hanxi Li\thanks{This work was conducted during Hanxi Li's visit to Zhejiang University.}}
\author[1,2,$\dagger$]{Zhengxun Zhang}

\author[2, $\star$]{Hao Chen}
\author[3, $\star$]{Lin Wu ~\IEEEmembership{Senior Member, ~IEEE}}
\author[4]{Bo Li}
\author[5]{Deyin Liu}
\author[1]{Mingwen Wang}

\affil[1]{Jiangxi Normal University, Jiangxi, China}
\affil[2]{Zhejiang University, Zhejiang, China}
\affil[3]{Department of Computer Science, Swansea University, SA1 8EN, United Kingdom}
\affil[4]{Northwestern Polytechnical University, Shannxi, China}
\affil[5]{Zhengzhou University, Henan, China}
\affil[$\dagger$]{These authors contributed equally to this work}
\affil[$\star$]{Corresponding author}

% The paper headers
\markboth{Journal of \LaTeX\ Class Files,~Vol.~14, No.~8, February~2024}%
{Shell \MakeLowercase{\textit{et al.}}: A Sample Article Using IEEEtran.cls for IEEE Journals}

\maketitle

\begin{abstract}
  Effectively addressing the challenge of industrial Anomaly Detection (AD) necessitates an ample supply of defective samples, a constraint often hindered by their scarcity in industrial contexts. This paper introduces a novel algorithm designed to augment defective samples, thereby enhancing AD performance. The proposed method tailors the blended latent diffusion model for defect sample generation, employing a diffusion model to generate defective samples in the latent space. A feature editing process, controlled by a ``trimap" mask and text prompts, refines the generated samples. The image generation inference process is structured into three stages: a free diffusion stage, an editing diffusion stage, and an online decoder adaptation stage. This sophisticated inference strategy yields high-quality synthetic defective samples with diverse pattern variations, leading to significantly improved AD accuracies based on the augmented training set. Specifically, on the widely recognized MVTec AD dataset, the proposed method elevates the state-of-the-art (SOTA) performance of AD with augmented data by 1.5\%, 1.9\%, and 3.1\% for AD metrics AP, IAP, and IAP90, respectively. The implementation code of this work can be found at the GitHub repository \url{https://github.com/GrandpaXun242/AdaBLDM.git}

\end{abstract}


\section{Introduction}
\label{sec:intro}

In a practical manufacturing workflow, obtaining defective samples is considerably more challenging compared to acquiring defect-free ones. Consequently, the majority of recently proposed algorithms for industrial defect inspection, such as \cite{CutPasteV2, defard_padim_2020, zavrtanik2021draem, PatchCore, yang2023memseg, zhang2023destseg, liu2023diversity, 9940966, 9910145, 10251020}, address the issue as an Anomaly Detection (AD) problem—an established machine learning challenge \cite{scholkopf2000support, chandola2009anomaly}. Among these industrial AD algorithms, the assumption is made that all anomaly-free samples, be they complete images or image patches, belong to a single distribution. Meanwhile, defective samples are identified as ``outliers". The strategic choice of adopting an Anomaly Detection approach for industrial defect inspection is evident: the absence of the need for defective samples during the training stage makes these AD-based algorithms inherently compatible with practical manufacturing scenarios.
% \cite{PatchCore, zhang_destseg_2023, li_target_2023, li_efficient_2023}. 

\begin{figure}[tp]
  \label{fig:three_method}
  \centering
  \includegraphics[scale=0.6]{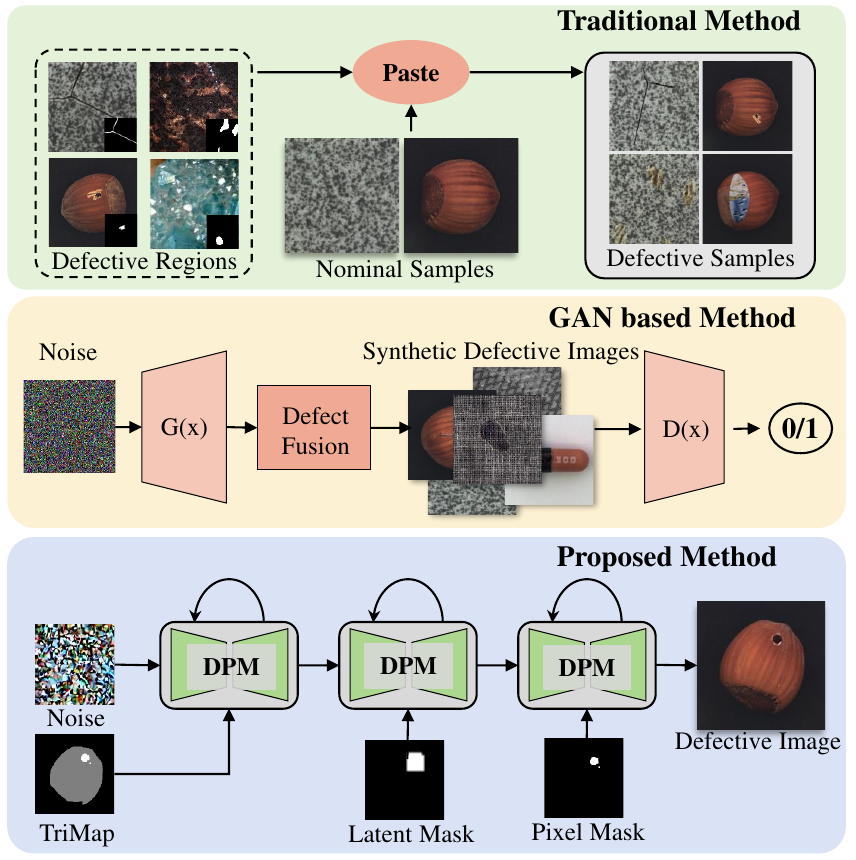}
  \caption{Illustration of three defect generation styles. From top to
  bottom: conventional approaches, GAN-based algorithms, and the proposed method.}
\end{figure}

Nonetheless, achieving this compatibility comes at the expense of an extremely imbalanced data distribution, a characteristic that is deemed unacceptable in the context of most discriminative algorithms. Consequently, some researchers suggest the generation of artificial defect patterns using conventional approaches \cite{CutPasteV2, zavrtanik2021draem, yang2023memseg} to facilitate subsequent discriminative learning processes. Recently, state-of-the-art (SOTA) algorithms \cite{huang_prototype-based_2023, li_target_2023, li2023efficient} suggest that incorporating a few real anomalous samples during the training stage can significantly enhance the AD accuracy if these samples are combined with synthetic samples. For example, some pioneering works \cite{WACVDCGAN, AAAI2023DFM}  demonstrated the benefits of generating these "lifelike" defects using more sophisticated methods. The first two rows of Fig.~\ref{fig:three_method} illustrate two different types of defect generation.

In this paper, we propose the integration of the cutting-edge AI generation algorithm, namely the Diffusion Probabilistic Model (DPM) \cite{sohl2015deep}, into the realm of defect generation. Specifically, we leverage the Blended Latent Diffusion Model \cite{LDM, avrahami2022blended, avrahami_blended_2023} to create our framework. To enhance the generation of defective samples, we innovate the  Blended Latent Diffusion Model (BLDM)\cite{avrahami_blended_2023} with three modules. First, we design a novel ``defect trimap" to delineate the target object mask and the defective regions on the generated image, incorporating it as a new form of controlling information alongside the language prompts within the stable diffusion model. Second, to ensure the authenticity of the generated samples, a cascaded ``editing" stage is introduced into the BLDM, at both the latent and pixel space. Finally, an innovative online adaptation of the image encoder is proposed to refine the quality of the generated images. Consequently, the customized BLDM algorithm, referred to as AdaBLDM in this paper, is capable of generating higher-quality defective samples with pattern variations. The third row of Fig.~\ref{fig:three_method} provides a brief illustration of the proposed BLDM-based method. In summary, the contributions of this paper are threefold, as outlined below.

\begin{itemize}
  \item
    To the best of our knowledge, it is the first time that the Diffusion Model has been adopted
    for the task of industrial defect generation. The advantages of the BLDM such as the
    generation stability, the high image quality, and the controllable content could all
    contribute to achieving a better defect generator compared with the SOTA methods. 
  \item
    To tailor the BLDM algorithm to the defect generation scenario, we bake BLDM in the aspects of controlling information, preheating stage, and
    online adaptation, respectively. This endeavor leads to the proposed algorithm, termed AdaBLDM, which achieves notably improved defective samples compared to synthetic ones generated by existing SOTA methods. 
  \item
    Given the defective samples generated by the proposed method, we
    achieved new SOTA performance  on the MVTec AD dataset \cite{bergmann2019mvtec}. In specific, the proposed AdaBLDM outperforms the best-to-date method DeSTseg \cite{zhang2023destseg}  by $\boldsymbol{1.5\%}$, $\boldsymbol{1.9\%}$,
    $\boldsymbol{3.1\%}$ on the AD metrics AP, IAP and IAP$\bf 90$, respectively.
\end{itemize}

The remainder of this paper is organized as follows. In Sec.~\ref{sec:related}, we discussed the related work in literature. The section \ref{sec:method} presents the
proposed method in detail. Experimental results and the ablation study are addressed in Sec.~\ref{sec:experiments}. We conclude this paper in Sec. \ref{sec:conclusion}.

\section{Related work}
\label{sec:related}

\subsection{Synthetic Defect Generation}
Due to the scarcity of defective samples, generating synthetic defective samples becomes a rule of thumb among researchers to achieve satisfactory performance \cite{CutPasteV2, defard_padim_2020, zavrtanik2021draem, yang2023memseg,
zhang2023destseg, LipFormer}. A straightforward way akin to data augmentation is mimicking the defects by ``pasting'' anomalous pixels on the normal
images. For example, methods in the regime
\cite{devries2017improved, CutPaste, CutPasteV2} randomly cut
regions from normal images and paste them to the ``incorrect'' places as artificial
defects. Crop\&Paste \cite{9428468} and PRN\cite{zhang2023prototypical} crop defect areas
from genuine defective images and paste them onto defect-free images. Those algorithms,
though achieve better AD performance compared with the standard one-class AD approaches,
can not create new defect patterns and thus likely lead to overfitting problems.  To
increase the variations of anomalies, DRAEM \cite{zavrtanik2021draem}, DeSTseg
\cite{zhang2023destseg}, MegSeg \cite{yang2023memseg} and ReSynthDetect
\cite{Niu_2023_BMVC} employ extra datasets combined with Berlin noise to fabricate
defects. However, in these cases, the distribution of fabricated defects differs from the
real one and thus the performance gain can not be ensured.

Inspired by the potent AIGC (AI Generated Contents) methods such as Generative Adversarial Network (GAN) based algorithms \cite{xia2022gan,niu2020defect,Cross-Entropy-TCSVT} and diffusion model-based algorithms
\cite{yang2023diffusion}, some attempts have been devoted to generating more natural-looking simulated defects. In particular, SDGAN \cite{niu2020defect} employs two
generators to switch the defective and non-defective status for sample images. As a result, SDGAN \cite{niu2020defect} can
generate a high-quality, diverse dataset of steel surfaces with both defective and
defect-free images. In a similar vein to SDGAN \cite{niu2020defect},
Defect-GAN \cite{zhang2021defect} converts non-defective images into defective ones to
enhance the learning process of the defect classifier. However, the above two GAN-based
generation algorithms cannot generate defective regions that are aligned precisely as required. Despite that they
could enhance the performance of defect classification, they cannot lift the defect segmentation accuracy which is crucial in many real-world applications.

More recently, \cite{WACVDCGAN} and \cite{AAAI2023DFM} proposed to generate both synthetic defective images and the
corresponding pixel-level labels. In specific, the DCDGANc algorithm
\cite{WACVDCGAN} designed a novel GAN-based method to mimic real defects, which can be
fused with defect-free images via an improved Poisson blending algorithm. At the same time, \cite{AAAI2023DFM} introduced DFMGAN, stemming from 
StyleGANv2 \cite{Karras2019AnalyzingAI}, to generate defective images and defective masks by
using the proposed defect-aware residual blocks. 
% Since DFMGAN relies on
% editing StyleGanV2 image features to obtain defect-free images, defect images, and defect
% masks, there exists discontinuity in feature editing space, 
Although the above two GAN-based methods can generate high-quality images with defects, they fall short in generating the defective mass strictly aligned to the generated defective pixels on the image. Such misalignment is ascribed to the disparity between the image-specific generative methods and the challenging scenario of AD, and thus significantly misleads the defective segmentation model in AD. In contrast, in this paper, we propose a novel defect generation method based on Diffusion Models. We
tame this general-purpose algorithm to generate diverse, realistic but accurately-controlled defective
image samples via a series of novel modifications.

\subsection{Diffusion Probabilistic Models for Image Editing}
In the last few years, researchers have proposed various approaches to realize
high-quality image generation. The off-the-shelf tools include Generative Adversarial
Networks (GANs) \cite{goodfellow2014generative}, Variational AutoEncoders (VAEs)
\cite{kingma2013auto}, the flow-based algorithms \cite{dinh2014nice}, and Diffusion
Probabilistic Models (DPMs) \cite{sohl2015deep, ho2020denoising, song2020denoising,
dhariwal2021diffusion}. In specific, DPMs have demonstrated higher qualities of the
synthetic images and impressive stability of training. Accordingly, the DPM-based
approaches, such as Stable Diffusion (Stability AI) \cite{LDM}, DALL-E2 (OpenAI)
\cite{ramesh2022hierarchical}, and Imagen (Google) \cite{saharia2022photorealistic}, have
achieved state-of-the-art performances in the task of content generation. 

DPM-based methods also play an important role in the remit of image editing. In particular, the Latent Diffusion Model \cite{LDM} conducts the image generation
or editing in the lower-dimensional latent space and thus can achieve higher efficiency; RePaint \cite{lugmayr_repaint_2022} fully leverages a pre-trained DDPM \cite{song2020denoising} to perform image editing by sampling given pixels during the reverse diffusion stage; DCFace \cite{kim2023dcface} employs two feature encoders to control the Diffusion Model to facilitate facial image editing; \cite{zhang_adding_2023} proposes the ``ControlNet'' to govern existing Diffusion models to achieve a better-controlled synthesis;  
\cite{avrahami2022blended, avrahami_blended_2023} enable image editing
by making modifications in either the pixel space or feature space; \cite{10154005} propose the "Dual-Cycle Diffusion" to generate an unbiased mask to guide image editing. While these methods demonstrate state-of-the-art (SOTA) performances in general image editing tasks, they can hardly produce image regions that are precisely aligned at the pixel level due to the inherently fuzzy nature of the reverse diffusion process. This nuisance can potentially decrease the segmentation capacity in anomaly detection. 

In this paper, we customize the cutting-edge Blended Latent Diffusion Model (BLDM) \cite{avrahami_blended_2023} to generate precisely aligned synthetic regions for industrial defect detection and localization. It is important to note that this work employs the diffusion model for generating synthetic images or image regions, just similar to the pioneering work \cite{sohl2015deep, wu2024datasetdm} proposed for the general editing tasks. Our method is irrelevant to the algorithms that use DPM to directly predict anomaly regions on the test images \cite{wyatt2022anoddpm, zhang2023unsupervised, wolleb2022diffusion, xu2023unsupervised}.   

% \subsection{Defect Inspection}
% In industrial manufacturing scenarios, scarcity of defective image data renders many fully
% supervised methods ineffective (due to overfitting issues). Employing a reconstruction
% strategy, DRAEM\cite{zavrtanik2021draem} utilizes restoration networks for anomaly
% detection. Employing a distillation strategy, DeSTseg\cite{zhang2023destseg} uses
% distilled denoising networks to infer defects when significant differences emerge between
% comparative data. These methods introduce additional texture data (DTD) to artificially
% create supervisory signals for anomaly detection tasks. By simulating defective images
% through a cascadinly blended generation strategy, \OUR  can generate defect samples that
% possess authenticity, flexibility, and accuracy.

\section{The proposed method}
\label{sec:method}

\subsection{Task Definition and Preliminaries}
\label{subsec:task}
Following the conventional setting of anomaly detection, an anomaly-free (OK) sample set
$\mathcal{X}_{OK} = \{\bb{x}^i_{OK} \in \hanxiR^{H_x \times W_x \times 3} \mid i = 1, 2,
  \dots, N_{OK}\}$ is available for training. Furthermore, as introduced in Sec.~\ref{sec:intro},
a few defective (NG) samples $\mathcal{X}_{NG} = \{\bb{x}^i_{NG} \in \hanxiR^{H_x \times W_x
    \times 3} \mid i = 1, 2, \dots, N_{NG}\}$ and the corresponding anomaly masks $\mathcal{M}_{NG}
  = \{\bb{M}^i_{NG} \in \mathbb{B}^{H_x \times W_x} \mid i = 1, 2, \dots, N_{NG}\}$ are given
as the ``seeds'' of defect generation and usually $N_{NG} \ll N_{OK}$. The goal of the proposed method is to generate a number of training samples with synthetic defective regions, \emph{i.e.}, 
\begin{equation}
  \label{equ:larger}
  \{\mathcal{X}_{OK}, \mathcal{X}_{NG}, \mathcal{M}_{NG}\}  \xrightarrow{\texttt{Defect
      Generator}} \{\mathcal{X}^{\ast}_{NG}, \mathcal{M}^{\ast}_{NG}\},
\end{equation}
where $\mathcal{X}^{\ast}_{NG} = \{\bb{x}^{\ast i}_{NG} \in \hanxiR^{H_x \times W_x \times
3} \mid i = 1, 2, \dots, N^{\ast}_{NG}\}$ and $\mathcal{M}^{\ast}_{NG} = \{\bb{M}^{\ast
i}_{NG} \in \mathbb{B}^{H_x \times W_x} \mid i = 1, 2, \dots, N^{\ast}_{NG}\}$ comprise
the obtained synthetic samples and the corresponding defect masks, respectively.
Furthermore, usually we set $N^{\ast}_{NG} \gg N_{NG}$ to
ensure high pattern variation of the generated samples.

In this paper, we propose to generate the synthetic samples using a diffusion model. It is common for diffusion models \cite{sohl2015deep, ho2020denoising} to approach generation target as a noise-free variable $\bb{x}_0$ and define the ``forward diffusion process'' as follows
\begin{equation}
  \label{equ:diffusion_forward}
  \begin{split}
     & q(\bb{x}_{t}\mid\bb{x}_{t-1}) =  \mathcal{N}(\bb{x}_t;\sqrt{1-\beta_t}
    \bb{x}_{t-1},\beta_t\II)                                                        \\
     & q(\bb{x}_{1:T}\mid\bb{x}_0) = \prod^{T}_{t =1}q(\bb{x}_{t}\mid\bb{x}_{t-1}),
  \end{split}
\end{equation}
where $\{\beta_t \in (0,1)\}^T_{t=1}$ denotes the predefined variance schedule and $\II$
is a identity matrix. When $T$ is large enough, the diffusion process could convert a
normal image into a random noise $\bb{x}_T$. On the contrary, the ``reverse diffusion
process'' is defined as
\begin{equation}
  \label{equ:diffusion_reverse}
  \begin{split}
     & p_{\theta}(\bb{x}_{t-1}\mid\bb{x}_{t}) = \mathcal{N}(\bb{x}_{t-1};
    \boldsymbol{\mu}_{\theta}(\bb{x}_t, t), \boldsymbol{\Sigma}_{\theta}(\bb{x}_t, t))               \\
     & p_{\theta}(\bb{x}_{0:T}) = p(\bb{x}_T)\prod^{T}_{t =1}p_{\theta}(\bb{x}_{t-1}\mid\bb{x}_{t}),
  \end{split}
\end{equation}
where $\boldsymbol{\mu}_{\theta}(\cdot)$ and $\boldsymbol{\Sigma}_{\theta}(\cdot)$ are the
two unknown functions that are usually approximated as deep models. In the test phase, given
different random Gaussian noise samples $\bb{x}^i_T, i = 1, \dots, N_{tst}$, one could
obtain high quality and various samples $\bb{x}^i_0, \forall i$ (usually RGB images) from
the same domain of the training samples. Akin to the above diffusion process, we could also generate lifelike anomalous images in
$\mathcal{X}^{\ast}_{NG}$ based on a diffusion model which has been well learned on the samples sets $\mathcal{X}_{OK}$ and $\mathcal{X}_{NG}$.

\subsection{Latent Diffusion Model and Cross-Modal Prompts}
\label{subsec:ldm}
\begin{figure*}[ht]
  \centering{
  \includegraphics[scale=0.45]{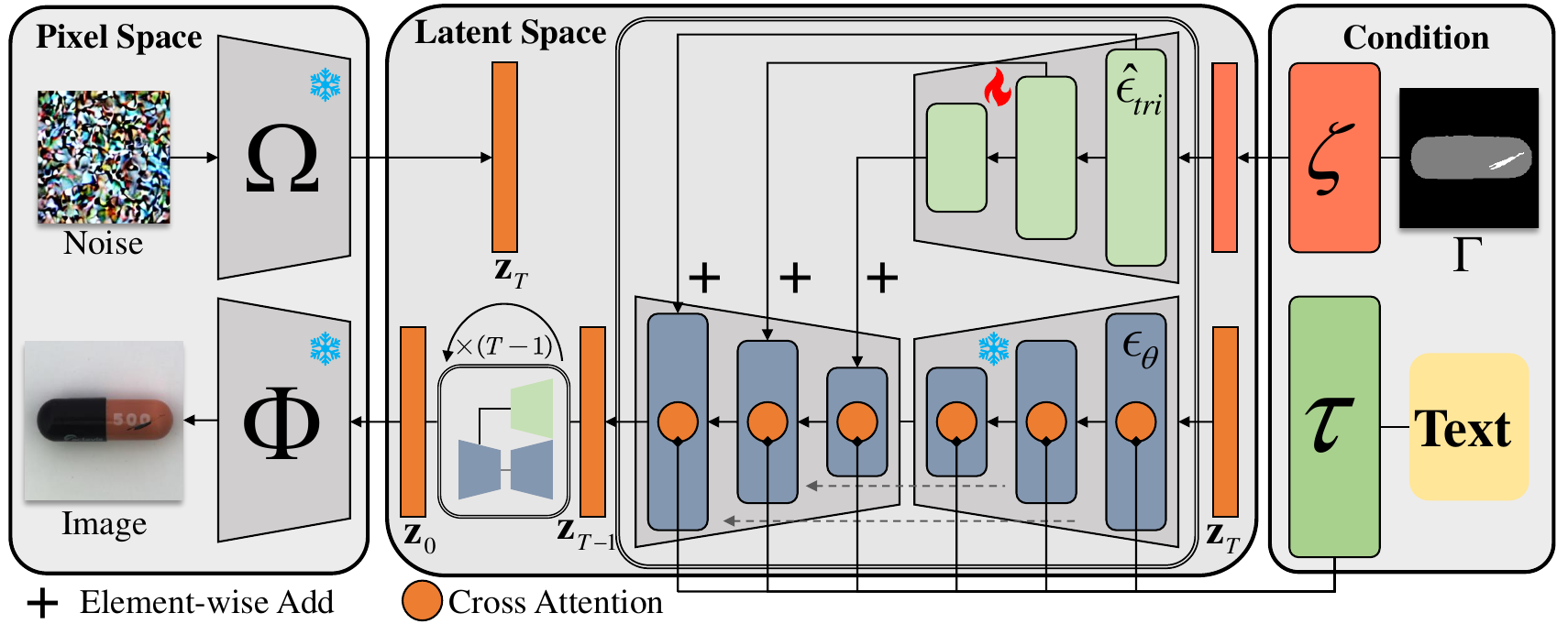}
  \caption{The network structure of the proposed BLDM-based method for generating defective regions on a image. One can see besides the noise inupt, the model is governed by a a text prompt and a trimap that indicates the locations of the object and defect}
  \label{fig:struct}
  }
\end{figure*}
\subsubsection{Latent diffusion model with controlling information}
In this paper, we employ the Latent Diffusion Model (LDM) as the backbone of our method
encouraged by its praiseworthy advantages \cite{LDM}. As the name indicating, the forward
and reverse diffusion processes of LDM are both performed in the lower-dimentional latent space.
Given an input image $\bb{x} \in \hanxiR^{H_x \times W_x \times 3}$, its
corresponding latent feature $\bb{z} \in \hanxiR^{H_z \times W_z \times C_z}$ is obtained
using the encoder function $\Omega(\cdot)$ as $\bb{z} = \Omega(\bb{x})$ while the image
can also be recovered from $\bb{z}$ via the decoding operation as $\bb{x} = \Phi(\bb{z})$. Based on the latent feature $\bb{z}$ and some simplifications \footnote{Readers are
referred to \cite{LDM} and \cite{ho2020denoising} for more details} from
Eq.~\ref{equ:diffusion_reverse}, the reverse conditional probability in latent space is revised as
\begin{equation}
  \label{equ:reverse_conditional}
  \begin{split}
    & p_{\theta}(\bb{z}_{t-1}\mid\bb{z}_{t}, t, \mathcal{C}) = \\
     & \mathcal{N}\left(\bb{z}_{t-1};
    \frac{1}{\sqrt{\alpha_t}}\left(\bb{z}_t - \frac{1 - \alpha_t}{\sqrt{1 -
    \bar{\alpha}_t}}\right){\epsilon}_{\theta}(\bb{z}_t, t, \mathcal{C}),
    \sigma_t\II\right),
  \end{split}
\end{equation}
where ${\epsilon}_{\theta}(\cdot)$ stands for the $\theta$-parameterized deep model to
be trained; $\mathcal{C}$ denotes the controlling prompts fed into LDM for stabilizing the
generation process \cite{LDM}; $\alpha_t$, $\bar{\alpha}_t$ and $\sigma_t$ the parameters
which can be calculated in a deterministic way at each $t$-th step. In our method, we
propose to generate cross-modal controlling features to enhance the quality of the synthetic
defect. 

\subsubsection{Linguistic prompts}
As an effective module, linguistic prompts are introduced to control and stabilize the generation process of diffusion models \cite{LDM, avrahami_blended_2023}. In this work, the linguistic prompt $y$ is obtained by specifying the keywords in the following template sentence
\begin{equation}
  y \coloneqq ``\{obj\},\text{a }\{obj\} \text{ with} \text{ a }\{def\}",
\end{equation}
where the keywords $obj$ and $def$ denote the category of the current object (as the capsule in
Fig.~\ref{fig:struct}) and defect type (as the black contamination on the capsule),
respectively. One can see this template design is straightforward and thus the prompt is
always easy to obtain. The sentence $y$ is then processed by the
parameter-frozen text encoder $\tau(\cdot): \hanxiR^{d_y} \rightarrow \hanxiR^{d_{lang}}$
of the CLIP model \cite{Radford2021LearningTV}. In particular, as
shown in Fig.~\ref{fig:struct}, the linguistic controlling information is fed into the
generation model $\epsilon_{\theta}(\cdot)$ and the generation process is governed via a 
K-Q-V attention mechanism proposed in \cite{LDM}. 

\subsubsection{Defect trimap prompts}
\label{subsubsec:trimap}
Recall that in practical anomaly detection tasks, the defect-free samples
are usually sufficient in terms of both amount and pattern variation, we thus propose to generate
defective samples based on defect-free prototypes, rather than generating them from scratch. In a
nutshell, a ``trimap'' $\Gamma \in \hanxiR^{H_x \times W_x}$ is designed to specify the
desired location of the generated object and the defect, as shown in Fig.~\ref{fig:struct}.
To obtain a proper trimap, we firstly estimate the foreground region \footnote{The
area corresponding to the interested object} on a randomly selected defect-free image $\bb{x}_{OK}$ by using the method proposed in \cite{li_target_2023} and save the
foreground mask as $\bb{F} \in \mathbb{B}^{H_x \times W_x}$ with $\bb{F}(x, y) = 1$
indicating a foreground pixel at the coordinate $[x, y]$. Secondly, a synthetic defect mask is created based on a genuine anomaly mask randomly selected from set $\mathcal{M}_{NG}$. Formally, the generation of the novel anomaly mask writes: 
\begin{equation}
  \label{equ:defect_mask}
  \mathcal{M}_{NG} \xrightarrow{\text{rand.} ~i} \bb{M}^i_{NG} \xrightarrow{\text{rand. affine}}
  \bb{M}^{A}_{NG} \xrightarrow{\text{fit}~\bb{F}} \bb{M}^{\ast}_{NG},
\end{equation}
where $\bb{M}^i_{NG}, \bb{M}^{A}_{NG}, \bb{M}^{\ast}_{NG} \in \mathbb{B}^{H_x \times
W_x}$; ``rand. $i$'' refers to a random index selection; ``rand. affine'' denotes a random
affine transform while ``fit $\bb{F}$'' stands for adjusting the location and scale of
$\bb{M}^{A}_{NG}$ so that the adjusted mask is entirely inside the foreground region
defined by $\bb{F}$. Finally, the value of each pixel on $\Gamma$ are defined as follows
\begin{equation}
  \label{equ:trimap}
  \Gamma(x, y) = 
  \begin{cases}
    1 \quad & \text{if}~\bb{M}^{\ast}_{NG}(x,y) = 1, \\
    0.5 \quad & \text{else if}~\bb{F}(x,y) = 1, \\
    0 \quad & \text{else}.
  \end{cases}
\end{equation}
% In this work, we empirically demonstrate this image-based controlling information can
% efficiently regularize the generation process of the latent diffusion model. 

As shown in Fig.~\ref{fig:struct}, in the proposed AdaBLDM algorithm, the controlling trimap
$\Gamma$ is firstly embedded by a convolutional block as $\zeta(\Gamma) \in
\hanxiR^{H_z \times W_z \times C_z}$. The embedding feature is then fed into an encoder network
$\hat{\epsilon}_{tri}(\cdot)$ that shares the same structure and initial parameters with
the denoising encoder of $\epsilon_{\theta}(\cdot)$, except that the original ``spatial
attention'' module is replaced with a self-attention process over the embedded trimap
features. The deep features extracted from different layers of $\hat{\epsilon}_{tri}(\cdot)$ are
then injected into the corresponding layers of the denoising decoder of $\epsilon_{\theta}(\cdot)$ via element-wise addition (see Fig.~\ref{fig:struct}). 

\subsection{Loss Function and Training Scheme}
\label{subsec:training}

Given the cross-modal controlling features $\{\tau(y), \zeta(\Gamma)\}$, the learning objective of the proposed model is defined as:
\begin{equation}
  \label{equ:LDM}
  \resizebox{.9\hsize}{!}{$L_{LDM} \coloneqq \mathbb{E}_{\bb{z}_t,\mathcal{C},\epsilon\thicksim
  \mathcal{N}(0,1),t}\left[\|\epsilon - \epsilon_{\theta}(\bb{z}_t, t,
    \tau(y), \hat{\epsilon}_{tri}(\zeta(\Gamma)))\|^2_2 \right],$}
\end{equation}
where ${\epsilon}_{\theta}(\cdot)$ denotes the denoising model and $\hat{\epsilon}_{tri}$ stands for the encoder for the latent trimap features. In this work, the involved modules are learned with different training schemes as described below. 
\begin{itemize}
    \item ${\epsilon}_{\theta}(\cdot)$ is  pre-trained by using the domain
            dataset (e.g., all training images of all the subcategories of MVTec AD
            \cite{bergmann2019mvtec} for the experiments on this dataset). We keep ${\epsilon}_{\theta}(\cdot)$ frozen during model learning for each subcategory.
    \item The trimap feature encoder
        $\hat{\epsilon}_{tri}(\cdot)$ is initialized by the corresponding parameters of the
        original LDM model \cite{LDM} and then fine-tuned for each subcategory based on the real
        defective samples
    \item The convolutional block $\zeta(\cdot)$ for trimap embedding is only learned based on the
        genuine anomaly samples from scratch. 
    % \item The text prompts $y$ are obtained by using the CLIP
    % \cite{Radford2021LearningTV} and BLIP \cite{Li2022BLIPBL} algorithm for sufficiently
    % various descriptions. 
    \item For the image encoder $\Omega(\cdot)$ and decoder $\Phi(\cdot)$, we follow \cite{LDM} to
        deploy the corresponding modules of the VQ-VAE algorithm \cite{van2017neural}, which are fixed during the training. \
    \item The text encoder $\tau(\cdot)$ is copied from CLIP and frozen during training. 
            
\end{itemize}

\subsection{Multi-Stage Denoising with Content Editing}
\label{subsec:reverse}

In this paper, we propose to perform feature editing in the inference phase by tailoring the editing scheme of Blended Latent Diffusion (BLD) \cite{avrahami2022blended} for defect generation. 

First
of all, given a defect-free image $\bb{x}_{OK}$ and a trimap $\Gamma$, the
corresponding defect mask $\bb{M}^{\ast}_{NG}$ is calculated as introduced in
Sec.~\ref{subsubsec:trimap}. Then the multi-stage denoising algorithm with content editing is
performed and the workflow of this algorithm is illustrated in Alg.~\ref{alg:editing}
where $\odot$ indicates element-wise multiplication and $\neg$ stands for the element-wise operation  of logical ``NOT'' on the binary map.

% Algorithm of Editing Diffusion Stage
\begin{algorithm}[t]
\label{alg:editing}
  \caption{Multi-Stage Denoising with Editing}
  \SetAlgoLined

  \KwIn{Defect-Free Image $\bb{x}_{OK}$, Trimap $\Gamma$, Defect Mask
  $\bb{M}^{\ast}_{NG}$, Language Prompt $y$, Decoder $\Psi(\cdot)$, Encoder
  $\Omega(\cdot)$, Range Numbers $T_1$, $T_2$ and $T_3$, Predefined Normal Distribution
  $\mathcal{N}(\mu_{z}, \sigma^2_z)$} 

  \KwOut{Blended Latent Feature $\bb{z}^{\ast}_{NG}$}

  $\bb{z}_{OK} \in \mathbb{R}^{H_z \times W_z \times C_z} \leftarrow
  \Omega(\bb{x}_{OK})$\\
  $\bb{M}^{z}_{NG} \in \mathbb{B}^{H_z \times W_z} \leftarrow Dilate \&
  DownSample(\bb{M}^{\ast}_{NG})$ \\
  $\bb{z}_{t} \leftarrow RandSapmle(\mathcal{N}(\mu_{z} , \sigma^2_z))$\\
  $t = T_1 + T_2 + T_3$ \\
  /*  \emph{Free Diffusion Stage}    */ \\
  \While{$t > T_2 + T_3$}{

    $\bb{z}_{t-1} \leftarrow Denoise(\bb{z}_t , \Gamma ,y)$ 
    \\
    $t = t-1$
  }
  /*  \emph{Latent Editing Stage}    */ \\
  \While{$t > T_3$}{

    $\bb{z}_t = \bb{z}_t \odot \bb{M}^z_{NG} + \bb{z}_{OK}\odot\neg \bb{M}^z_{NG}$ 

    $\bb{z}_{t-1} \leftarrow Denoise(\bb{z}_t , \Gamma , y)$\\
    $t = t-1$
  }
  /*  \emph{Image Editing Stage}    */ \\
  \While{$t >= 0$}{
    $\bb{x}_t = \Psi(\bb{z}_t)$\\
    $\bb{z}_t \leftarrow \Omega(\bb{x}_t \odot \bb{M}_{NG} + \bb{x}_{OK} \odot \neg \bb{M}_{NG})$\\
    $\bb{z}_{t-1} \leftarrow Denoise(\bb{z}_t,\Gamma,y)$\\
    $t = t-1$
  }
%   %\If{Category is Texture}{}	}
%   $Background Refine Step$

  \Return{$\bb{z}^{\ast}_{NG} = \bb{z}_0$}

\end{algorithm}

\begin{figure*}[tp]

  \centering{
  \includegraphics[scale=0.5]{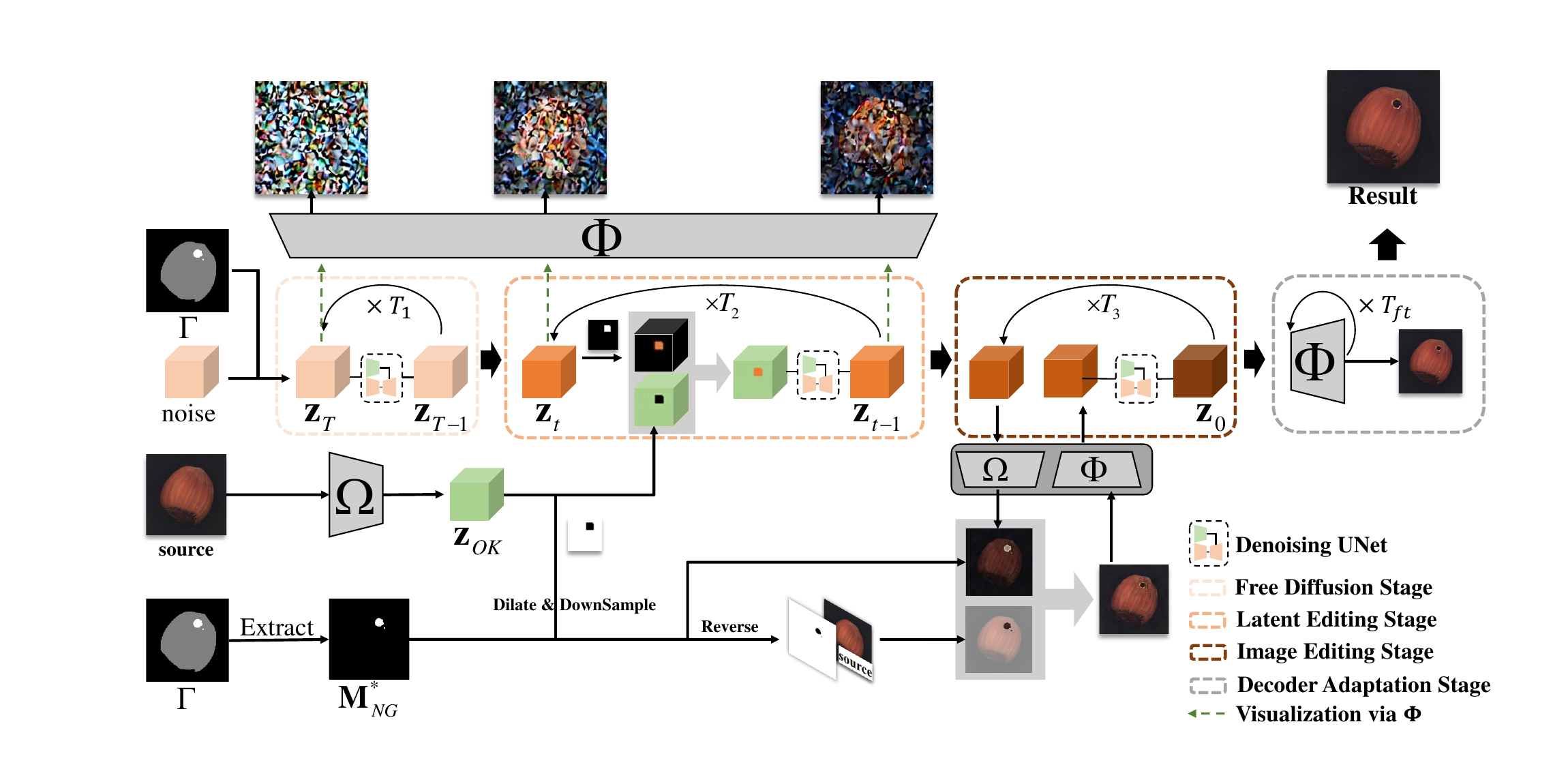}
  \caption{The inference scheme of the proposed AdaBLDM algorithm. One can see that the whole procedure can be mainly divided into $4$ stages, namely the pure denoising stage without editing; the latent editing stage; the image editing stage, and the decoder adaptation stage.}
    \label{fig:infer}}
\end{figure*}

As shown in Fig.~\ref{fig:infer}, the pipeline consists of three stages in the denoising phase. The first stage takes $T_1$ steps to conduct the conventional denosing\cite{ho2020denoising,song2020denoising,LDM} without content editing.
The second stage spends $T_2$ steps and the input feature $\bb{z}_t$ is merged with the
feature of $\bb{x}_{OK}$, under the guidance of the dilated defect mask
$\bb{M}^{\ast}_{NG}$. In the third stage, a novel content blending method proposed in this
paper is conducted. In specific, the blending operation is performed in the pixel space
and the blended image is then remapped back to the latent space for denoising. The image encoder $\Omega(\cdot)$ and decoder $\Phi(\cdot)$ are used cooperatively to switch the information between the two spaces. 

\subsection{Online Decoder Adaptation}
\label{subsec:online}
Considering that the image quality of the synthetic defective sample can be significantly influenced by the image decoder $\Phi(\cdot)$, we propose to fine-tune the decoder for each generated sample. Once the blended latent feature $\bb{z}^{\ast}_{NG} \in \hanxiR^{H_z \times W_z \times
C_z}$ is obtained from Alg.~\ref{alg:editing}, and given the defective mask $\bb{M}^{\ast}_{NG}
\in \mathbb{B}^{H_x \times W_x}$ and the source image $\bb{x}_{OK} \in \hanxiR^{H_x \times
W_x \times 3}$, we can fine-tune the decoder model $\Phi(\cdot)$ by conducting the online
update algorithm as described in Alg.~\ref{alg:online}, which shows that the online adaptation strategy encourages the decoder $\Phi(\cdot)$
generating defect-free pixels assembling to $\bb{x}_{OK}$ and simultaneously outputting the defective pixels guided by the original decoder. The conservative ratio $\lambda_{con}$ balances
the two objectives. In the experiments of this paper, we demonstrate that this online-updated decoder can generate more realistic samples for AD training and thus leads to higher AD performance. 

% Algorithm of Online Decoder Adaptation Stage
\begin{algorithm}[t]
  \label{alg:online}
  \caption{Online Decoder Adaptation}

  \SetAlgoLined
  \KwIn{Blended Latent Feature $\bb{z}^{\ast}_{NG}$, Defect-Free Image $\bb{x}_{OK}$,
  Maximum Step $T_{ft}$, Conservative Ratio $\lambda_{con}$, Anomaly Mask
  $\bb{M}^{\ast}_{NG}$, Decoder $\Phi(\cdot)$}

  \KwOut{Refined Image $\bb{x}^{\ast}_{NG}$}

  $\bb{x}^{\ast}_{NG} \leftarrow \Phi(\bb{z}^{\ast}_{NG}),~\bar{\bb{M}}^{\ast}_{NG}(i,j)
  \leftarrow \neg \bb{M}^{\ast}_{NG}(i,j)$ \\
  $t = 1$ \\
  \While{$t <= T_{ft}$}{

  $\tilde{\bb{x}} \leftarrow \Phi(\bb{z}^{\ast}_{NG})$\\

  $L_{i} = \sum_{i,j}^{H_x W_x} \left(\tilde{\bb{x}}(i,j) -
  \bb{x}_{OK}(i,j)\right)^2 \odot \bar{\bb{M}}^{\ast}_{NG}(i,j)$ \\

  $L_{d} = \sum_{i,j}^{H_x W_x} \left(\tilde{\bb{x}}(i,j) -
  \bb{x}^{\ast}_{NG}(i,j)\right)^2 \odot \bb{M}^{\ast}_{NG}(i,j)$ \\

  $\Phi(\cdot) \leftarrow AdamW(L_{i} + \lambda_{con} L_{d}, \Phi(\cdot))$\\

  $t = t + 1$

  }
  $\bb{x}^{\ast}_{NG} \leftarrow \Phi(\bb{z}^{\ast}_{NG})$    \\

  \Return{$\bb{x}^{\ast}_{NG}$}

\end{algorithm}

\subsection{Implementation Details}
\label{subsec:details}

In this paper, all the images are resized to $256 \times 256$ following the setting of
DeSTseg \cite{zhang2023destseg} which is a SOTA anomaly detection algorithm adopted by
this paper for evaluating the effectiveness of the generated samples. In this way, we set $H_x = W_x
= 256$ and $H_z = W_z = 32$. The latent feature channel number $C_z$ is $4$ in this paper
for efficient denoising operation. 
% Subsequently, we employ a  trained VQ-VAE \cite{van2017neural} with a pixel space
% $\mathbb{I} \subset 256 \times 256 \times 32$ to transform it into a latent space
% $\mathbb{Z} \subset 32 \times 32 \times 4$ for training the Blend Latent Diffusom model.
% During the Stable Diffusion Data Domain Finetune stage, we load the pretrained model named
% \emph{StableDiffusion-v1.5}. Regarding dataset construction, the image-text pairs for
% training (including defect-free data) are obtained using BLIP\cite{Li2022BLIPBL}  and
% CLIP\cite{Radford2021LearningTV}  to acquire corresponding descriptions for the data.
% These descriptions are  prepended with the object's name (ObjectName) to form the Text
% Prompt, denoted as $Prompt = \{Object Name, Description\}$.
As to the trimap prompts, the object's region is estimated by using the foreground
estimation method proposed in \cite{li_target_2023} and the anomalous mask is obtained
based on the randomly selected $10$ genuine defective samples for each subcategory of the
dataset. The affine transformation matrix mentioned in Sec.~\ref{subsubsec:trimap} is
determined by a random rotation angle $\gamma \in [0^{\circ}, 360^{\circ}]$ and a random
scaling factor $s \in [0.85, 1.15]$. For the AD tasks on texture subcategories, such as a
wooden or a fabric surface, all the pixels are considered as foreground except the
defective ones.

% The Training step can be referenced from Figure\ref{fig:struct}. The TextPrompt
% input\cite{jeong2023winclip} is structured as $Prompt = \{Object\},\text{a }\{Object\}
% \text{ with} \text{ a }\{Category\}$. Here, Object represents the name of the object
% (e.g., hazelnut, wood, etc.), and Category denotes the name of the object's defect (e.g.,
% crack, hole, etc.).  Text features are extracted using ClipTextEmbedder
% \cite{Radford2021LearningTV}, aligning with Domain Finetune.  In the original Denoising
% Unet, the parameters of Encoder $\mathbb{E}$ and Decoder $\mathbb{D}$ are no longer
% updated.  Before training AdaBLDM, Trimap Aware Control module initializes its parameters
% with $\mathbb{E}$. We cancel the spatial attention mechanism related to text and modifies
% it to use Trimap self-attention mechanism in this moudle. 
We employed the AdamW optimizer \cite{Loshchilov2017DecoupledWD} for updating the model
parameters, with a default learning rate of $1 \times 10^{-5}$, and default betas are set
to $0.9$ and $0.999$, respectively. During training, there is a $10\%$ chance that an empty
text prompt is input to the model. During the inference stage, we adopted the DDIM \cite{song2020denoising} sampling method
with the normal denoising step $T_1 = 50$, the latent blended step $T_2 = 30$, and pixel
blended step $T_3 = 5$. In the online decoder adaptation stage, $T_{ft} = 200$ and the
conservative ratio $\lambda_{con} = 100$ to avoid overfitting to the anomaly-free region.
The same optimizer as offline learning is used for this online adaptation except for the
learning rate is set to $1\times10^{-4}$.

\section{Experiments}
\label{sec:experiments}

\subsection{Experiment Setting}

\subsubsection{Evaluation Methodology}
In this section, we evaluate the proposed method for generating defective samples practically and extensively.  In most AIGC applications, the appearance quality of the
generated image, measured in either quantitative or qualitative ways, is the most
important criterion to evaluate a generation algorithm. However, in the scenario of
defective sample generation, we claim that a more important measuring standard is the
improvement in AD accuracy.  Consequently, a state-of-the-art anomaly detection
algorithm, namely DeSTseg \cite{zhang2023destseg} is learned based on the augmented
dataset, and the corresponding performance gain (if any) directly reflects the
effectiveness of the synthetic samples. In addition, the classical Support Vector Machine
(SVM) \cite{cortes1995support} is also employed as a complement to DeSTseg. Specifically,
the SVM model is learned to classify the pixels into normal and anomalous, and in this way,
a vanilla anomaly detection/localization method is obtained. The combination of the
sophisticated (DeSTseg) and the naive (SVM) AD algorithms increases the comprehensiveness
and thus the reliability of our experiment.

\subsubsection{Competitors}
We compare our defect generation algorithm with the cutting-edge generative models
including DFM \cite{AAAI2023DFM} and DCDGANc \cite{WACVDCGAN} which is built based on
StarGAN \cite{choi2020StarGAN} or StyleGANv2 \cite{Karras2019AnalyzingAI} algorithms. However, as the original StyelGANv2 in DCDGANc can easily lead to training crash according to our experiment, in this work, the StyleGANv2 is replaced by StyleGANv2-Ada \cite{NEURIPS2020_8d30aa96} to ensure the stability of training.    
Furthermore, the results of the original DeSTseg algorithm, which also generates synthetic defects in the pasting fashion, are also compared. The training samples of the
compared methods are strictly identical to those of AdaBLDM.

\subsubsection{Datasets}
To achieve a more general study, we perform the comparison on three well-known datasets,
namely MVTEC-AD \cite{bergmann2019mvtec} ($7$ subcategories over $15$), BTAD
\cite{mishra2021btad} (full dataset), and KSDD2 \cite{bovzivc2021mixed} which are
introduced below. 
\begin{itemize}
  \item \textbf{MVTec AD}\cite{bergmann2019mvtec} is an open dataset containing ten object
    categories and five texture categories commonly seen, with up to eight defect
    categories for each object/texture category. All the images are accompanied by
    pixel-level masks showing the defect regions. Among the available defect categories,
    we opted to utilize defect categories that contain more than $10$ defect images as the
    training dataset for our model.  Specifically, we randomly selected $10$ defect images
    per defect category from the test set and total defect-free images from the train set for
    model training and defect data generation. Due to time constraints, within the MVTec AD
    dataset, we focused on evaluating our model using data from $2$ Object categories
    (Hazelnut and Capsule) and $5$ Texture categories.
  \item \textbf{BTAD} \cite{mishra2021btad} comprises $2540$ images from three categories
    of real-world industrial products with different body and surface defects.  It is
    usually considered as a complementary dataset to the MVTec AD when evaluating an AD
    algorithm. Consistent with MVTec AD, we similarly opt for the selection of 10 instances
    of defect data along with defect-free data for the generation and training of the
    model.
  \item \textbf{KSDD2} \cite{bovzivc2021mixed} with over $3000$ images obtained while
    addressing a real-world industrial problem. This dataset comprises only one category,
    focusing on surface defects. For model training and data generation, we exclusively
    utilize the defect data and defect-free data from the training set.
\end{itemize}

To the best of our knowledge, the experiment of this paper involves much more subcategories, both in terms of number and variations, than any others in the current
literature.

\subsubsection{Evaluation metrics for Anomaly Detection and Localization}
\begin{itemize}
  \item \textbf{Pixel-AUC} is the area under the receiver operating characteristic curve
    at the pixel level. It is the most popular AD measuring method while failing to reflect
    the real performance difference between algorithms when a serious class imbalance
    exists.
  \item \textbf{PRO} \cite{pro} on the contrary, focuses on the anomaly pixels and treats
    the AD performance on each anomaly region equally. Consequently, the PRO
    metric is more robust to the class imbalance which is a common situation in
    most AD benchmarks.
  \item \textbf{AP} \cite{saito2015precision} metric as a conventional metric for semantic
    segmentation, is frequently adopted in recently proposed AD algorithms. It reflects the
    anomaly detection performance from a pixel-level perspective.
  \item \textbf{IAP} \cite{zhang2023destseg} focus on instance recall as a more
    straightforward metric.
  \item \textbf{IAP90} For those applications requiring an extremely high recall, the
    precision at recall = k\% is also computed and denoted as IAP@k, and k we follow
    \cite{zhang2023destseg} by setting k = $90$ in our experiments.
\end{itemize}

\subsubsection{Evaluation metrics for generation quality}
Following the conventional evaluation manner for image generation algorithms, we here also
compare the scores related to the quality of the generated images. In particular, the
\textbf{Kernel Inception Distance (KID)} \cite{binkowski2018demystifying} as well as the \textbf{Learned
Perceptual Image Patch Similarity (LPIPS)} \cite{zhang2018unreasonable} are involved in the
comparison. In simple words, the former criterion reflects the generation's authenticity while the
latter one indicates the generation's diversity.  

\subsubsection{Training Settings for the Involved AD Models}

In our experiments, the generated samples are evaluated by using the simplest AD
method and a sophisticated AD method. 
As one end of the spectrum, the SVM-based anomaly detection algorithm is extremely simple
and thus can directly reflect the primitive quality of the generated images, at the patch level. In specific,
the SVM model based on the RBF kernel \cite{chung2003radius} is employed as the patch
classifier. The patch features are extracted by using the PatchCore algorithm
\cite{PatchCore} and each $768$-D feature is assigned a positive label if the corresponding
patch is normal while negative labels are assigned to the anomalous ones. In each trial,
$100$ images are randomly sampled from the synthetic defective images and then $5000$
positive patch features and $5000$ negative patch features are extracted for training the
SVM model.  The final performance on each subcategory is calculated by averaging the
results of $10$ random trails with the same setting. Besides, the performance of the SVM
model trained merely on the genuine defective data is also reported. As to the parameter setting, we by default set the $\gamma$ of the RBF kernel as $1\times 10^{-4}$ while the regularization parameter $C$ is set to $1$. 

Referring to DeSTseg \cite{zhang2023destseg} which employs a
discriminative model to detect anomalous regions by comparing them with reconstructed defect-free counterparts, before training, we first generate
$5000$ defective samples by using AdaBLDM and another $5000$ defective samples by using
the data augmentation method of DeSTseg. We store the defective samples generated by
AdaBLDM in $\mathcal{D}_{gen}$ and store those augmented by DeSTseg in
$\mathcal{D}_{raw}$. The training data is then sampled from the defect-free samples,
$\mathcal{D}_{raw}$ and $\mathcal{D}_{gen}$ in the ratio of $5 : 4 : 1$. Note that
according to the methodology of DeSTseg, the genuine defective samples, which are the
seeds of the synthetic data generation, can not be used for training as the anomaly-free
version of them is unknown. For each subcategory, the DeSTseg is learned for $10000$ iterations and we pick the DeSTseg model with a UNIQUE iteration number for all the subcategories in one dataset.   

\subsection{Quantitative Results}
\subsubsection{Results on MVTec AD}
We first conduct the comparison of the SVM-based anomaly detection on the well-known
MVTec AD dataset and report the results in Tab.~\ref{tab:svm_mvtec}. From the table, we can
see that the SVM achieves consistently best accuracies based on the samples generated by
AdaBLDM. The genuine samples achieve the second-best performance while the samples
obtained by comparing SOTA generation algorithms perform worse than the two
aforementioned competitors.  
% MVtec Result
% SVM mvtec
\begin{table*}[h!]
\label{tab:svm_mvtec}
  \begin{center}
    \caption{The Anomaly Detection and Localization Performance of SVM on the
    MVTec AD dataset, Pixel-Auc/PRO/AP/IAP/IAP90, the best and the second-best numbers are
    shown in red and blue, respectively.}
    \resizebox{\linewidth}{!}{
      \begin{tabular}{c c c c c c c}
        \toprule
        \textbf{Category} & \textbf{Genuine}& \makecell{\textbf{DFM}\cite{AAAI2023DFM} \NL (AAAI2023)} & \makecell{\textbf{DCDGANc-StarGAN}\cite{WACVDCGAN} \NL (CVPRW2023)}& \makecell{\textbf{DCDGANc-StyleGAN}\cite{WACVDCGAN}\NL (CVPRW2023)} & \makecell{\textbf{\OUR} \NL (ours) }\\
        
        \hline
        Hazelnut          & \underline{\textcolor{blue}{98.07}}/\underline{\textcolor{blue}{70.03}}/\underline{\textcolor{blue}{47.07}}/\underline{\textcolor{blue}{94.85}}/\underline{\textcolor{blue}{63.03}} & 97.65/66.05/38.7/93.87/57.6  & 82.41/54.24/19.19/71.69/52.3                                                                                                                                                & 96.79/67.63/41.27/94.18/59.98                                                                                                                                           & \textbf{\textcolor{red}{98.23}}/\textbf{\textcolor{red}{71.08}}/\textbf{\textcolor{red}{47.94}}/\textbf{\textcolor{red}{94.88}}/\textbf{\textcolor{red}{66.65}}                     \\
        Wood              & 95.22/63.53/36.95/91.22/55.33                                                                                                                                                       & 39.91/3.9/2.89/12.99/5.12    & \textbf{\textcolor{red}{96.79}}/\underline{\textcolor{blue}{67.63}}/\underline{\textcolor{blue}{41.27}}/\textbf{\textcolor{red}{94.18}}/\underline{\textcolor{blue}{59.98}} & 89.83/51.97/25.08/81.6/44.96                                                                                                                                            & \underline{\textcolor{blue}{95.96}}/\textbf{\textcolor{red}{71.02}}/\textbf{\textcolor{red}{48.83}}/\underline{\textcolor{blue}{93.05}}/\textbf{\textcolor{red}{63.41}}             \\
        Capsule           & 92.28/13.91/6.65/84.01/12.31                                                                                                                                                        & 91.72/37.37/9.01/87.36/25.69 & \textbf{\textcolor{red}{99.51}}/\textbf{\textcolor{red}{71.09}}/\textbf{\textcolor{red}{52.11}}/\textbf{\textcolor{red}{99.14}}/\textbf{\textcolor{red}{61.98}}             & 94.05/37.13/10.47/89.09/28.15                                                                                                                                           & \underline{\textcolor{blue}{94.17}}/\underline{\textcolor{blue}{45.26}}/\underline{\textcolor{blue}{13.12}}/\underline{\textcolor{blue}{91.77}}/\underline{\textcolor{blue}{32.86}} \\
        Leather           & 99.07/62.88/40.86/98.52/51.21                                                                                                                                                       & 45.15/3.05/0.66/29.62/2.25   & 89.83/51.97/25.08/81.6/44.96                                                                                                                                                & \textbf{\textcolor{red}{99.51}}/\textbf{\textcolor{red}{71.09}}/\underline{\textcolor{blue}{52.11}}/\underline{\textcolor{blue}{99.14}}/\textbf{\textcolor{red}{61.98}} & \underline{\textcolor{blue}{99.23}}/\underline{\textcolor{blue}{68.5}}/\textbf{\textcolor{red}{60.52}}/\textbf{\textcolor{red}{99.56}}/\underline{\textcolor{blue}{61.49}}          \\
        Grid              & 89.02/28.68/2.87/75.29/23.78                                                                                                                                                        & 73.75/3.53/0.78/47.6/3.7     & 93.66/\underline{\textcolor{blue}{46.29}}/\textbf{\textcolor{red}{19.09}}/83.09/\textbf{\textcolor{red}{45.71}}                                                             & \textbf{\textcolor{red}{95.82}}/\textbf{\textcolor{red}{46.91}}/\underline{\textcolor{blue}{12.81}}/\textbf{\textcolor{red}{92.49}}/\underline{\textcolor{blue}{42.47}} & \underline{\textcolor{blue}{95.22}}/35.31/7.55/\underline{\textcolor{blue}{89.01}}/32.86                                                                                            \\
        Tile              & \underline{\textcolor{blue}{93.94}}/\textbf{\textcolor{red}{74.81}}/\textbf{\textcolor{red}{49.55}}/\underline{\textcolor{blue}{87.72}}/\textbf{\textcolor{red}{68.71}}             & 73.54/38.69/7.85/50.4/43.48  & \textbf{\textcolor{red}{94.05}}/37.13/10.47/\textbf{\textcolor{red}{89.09}}/28.15                                                                                           & 82.41/54.24/19.19/71.69/52.3                                                                                                                                            & 92.74/\underline{\textcolor{blue}{71.86}}/\underline{\textcolor{blue}{47.4}}/85.9/\underline{\textcolor{blue}{68.46}}                                                               \\
        Carpet            & 95.22/\underline{\textcolor{blue}{63.53}}/\underline{\textcolor{blue}{36.95}}/91.22/\underline{\textcolor{blue}{55.33}}                                                             & 39.91/3.9/2.89/12.99/5.12    & \underline{\textcolor{blue}{95.54}}/44.8/13.12/\underline{\textcolor{blue}{91.49}}/40.75                                                                                    & 89.83/51.97/25.08/81.6/44.96                                                                                                                                            & \textbf{\textcolor{red}{95.96}}/\textbf{\textcolor{red}{71.02}}/\textbf{\textcolor{red}{48.83}}/\textbf{\textcolor{red}{93.05}}/\textbf{\textcolor{red}{63.41}}                     \\
        \textbf{Average}  & \underline{\textcolor{blue}{95.06}}/\underline{\textcolor{blue}{54.48}}/\underline{\textcolor{blue}{32.97}}/\underline{\textcolor{blue}{89.49}}/\underline{\textcolor{blue}{48.11}} & 72.09/24.93/9.76/54.7/22.52  & 93.11/53.31/25.76/87.18/47.69                                                                                                                                               & 93.23/54.0/26.34/87.62/48.1                                                                                                                                             & \textbf{\textcolor{red}{95.95}}/\textbf{\textcolor{red}{57.69}}/\textbf{\textcolor{red}{35.51}}/\textbf{\textcolor{red}{91.57}}/\textbf{\textcolor{red}{52.53}}                     \\
        \bottomrule
      \end{tabular}
    }
  \end{center}
\end{table*}

On the other hand, Tab.~\ref{tab:destseg_mvtec} illustrates the AD performances of DeSTseg
based on the MVTec AD dataset. As shown in the table, the proposed method ranks the
first for all the $5$ AD metrics. In particular, the performance gains brought by the synthetic
samples generated by AdaBLDM are $1.00\%$, $1.98\%$ $6.68\%$, $6.35\%$ and $7.93\%$ on
Pixel-AUC, PRO, AP, IAP, and IAP90, respectively. The significant improvements prove the
validness of the proposed method.  
\begin{table*}[h!]
\label{tab:destseg_mvtec}
  \begin{center}
    \caption{The Anomaly Detection and Localization Performance of DeSTseg on the
    MVTec AD dataset, Pixel-Auc/PRO/AP/IAP/IAP90, the best and the second-best numbers are
    shown in red and blue, respectively.}
    \resizebox{\linewidth}{!}{
      \begin{tabular}{c c c c c c c}
        \toprule
      \textbf{Category} & \makecell{\textbf{DESTSEG}\cite{zhang2023destseg}\NL (CVPR2023)}& \makecell{\textbf{DFM}\cite{AAAI2023DFM} \NL (AAAI2023)}& \makecell{\textbf{DCDGANc-StarGAN}\cite{WACVDCGAN} \NL(CVPRW2023)}&\makecell{\textbf{DCDGANc-StyleGAN}\cite{WACVDCGAN}\NL (CVPRW2023)}& \makecell{\textbf{\OUR} \NL (ours) }\\
        \hline
        Hazelnut          & \textbf{\textcolor{red}{99.65}}/97.46/88.22/89.06/75.58                                                            & 99.45/98.01/88.5/90.48/\textbf{\textcolor{red}{82.32}}                                   & 99.53/97.69/87.47/87.78/74.17                                                                                                                                       & 99.62/\underline{\textcolor{blue}{98.1}}/\underline{\textcolor{blue}{90.31}}/\underline{\textcolor{blue}{90.69}}/80.78                                                             & \textbf{\textcolor{red}{99.65}}/\textbf{\textcolor{red}{98.65}}/\textbf{\textcolor{red}{91.27}}/\textbf{\textcolor{red}{91.47}}/\underline{\textcolor{blue}{81.96}} \\
        Wood              & 95.72/93.0/73.62/79.77/63.6                                                                                        & 97.91/97.09/83.78/\underline{\textcolor{blue}{92.49}}/\textbf{\textcolor{red}{85.63}}    & 97.83/96.99/84.14/\textbf{\textcolor{red}{92.63}}/85.05                                                                                                             & \textbf{\textcolor{red}{98.34}}/\underline{\textcolor{blue}{97.17}}/\textbf{\textcolor{red}{84.67}}/91.98/\underline{\textcolor{blue}{85.24}}                                      & \underline{\textcolor{blue}{98.23}}/\textbf{\textcolor{red}{97.23}}/\underline{\textcolor{blue}{84.59}}/90.85/82.47                                                 \\
        Capsule           & \textbf{\textcolor{red}{99.21}}/\textbf{\textcolor{red}{96.29}}/57.79/52.7/33.32                                   & 98.43/95.39/\underline{\textcolor{blue}{62.89}}/\underline{\textcolor{blue}{58.5}}/31.82 & 98.44/94.78/61.07/56.63/30.49                                                                                                                                       & \underline{\textcolor{blue}{98.85}}/95.77/62.08/57.28/\underline{\textcolor{blue}{34.16}}                                                                                          & 98.61/\underline{\textcolor{blue}{96.14}}/\textbf{\textcolor{red}{63.39}}/\textbf{\textcolor{red}{61.04}}/\textbf{\textcolor{red}{39.0}}                            \\
        Leather           & 99.76/99.06/75.0/74.96/64.92                                                                                       & 99.81/99.42/80.08/83.28/72.91                                                            & \textbf{\textcolor{red}{99.84}}/\textbf{\textcolor{red}{99.56}}/\underline{\textcolor{blue}{82.53}}/\textbf{\textcolor{red}{86.47}}/\textbf{\textcolor{red}{76.82}} & \textbf{\textcolor{red}{99.84}}/99.5/81.7/83.64/\underline{\textcolor{blue}{74.32}}                                                                                                & \textbf{\textcolor{red}{99.84}}/\textbf{\textcolor{red}{99.56}}/\textbf{\textcolor{red}{82.62}}/\underline{\textcolor{blue}{85.89}}/73.51                           \\
        Grid              & \underline{\textcolor{blue}{98.82}}/94.3/\underline{\textcolor{blue}{57.74}}/57.11/\textbf{\textcolor{red}{35.25}} & 98.24/\underline{\textcolor{blue}{94.38}}/56.0/57.38/26.34                               & 98.21/93.52/57.2/\underline{\textcolor{blue}{57.52}}/20.91                                                                                                          & 97.88/91.47/56.4/55.95/13.72                                                                                                                                                       & \textbf{\textcolor{red}{98.95}}/\textbf{\textcolor{red}{96.21}}/\textbf{\textcolor{red}{60.65}}/\textbf{\textcolor{red}{62.87}}/\underline{\textcolor{blue}{31.11}} \\
        Tile              & 98.7/97.28/93.12/96.38/\textbf{\textcolor{red}{92.33}}                                                             & 98.68/96.77/92.87/95.58/88.65                                                            & 99.04/97.54/\underline{\textcolor{blue}{94.1}}/\underline{\textcolor{blue}{96.41}}/\underline{\textcolor{blue}{90.38}}                                              & \underline{\textcolor{blue}{99.1}}/\underline{\textcolor{blue}{97.69}}/93.97/96.27/90.24                                                                                           & \textbf{\textcolor{red}{99.17}}/\textbf{\textcolor{red}{97.86}}/\textbf{\textcolor{red}{94.57}}/\textbf{\textcolor{red}{96.61}}/90.14                               \\
        Carpet            & 94.86/92.99/69.74/84.79/58.4                                                                                       & 97.89/96.64/78.99/88.6/74.26                                                             & 98.18/97.08/81.45/89.28/76.53                                                                                                                                       & \underline{\textcolor{blue}{98.55}}/\underline{\textcolor{blue}{97.5}}/\underline{\textcolor{blue}{82.07}}/\underline{\textcolor{blue}{90.34}}/\underline{\textcolor{blue}{78.81}} & \textbf{\textcolor{red}{99.33}}/\textbf{\textcolor{red}{98.47}}/\textbf{\textcolor{red}{84.87}}/\textbf{\textcolor{red}{90.53}}/\textbf{\textcolor{red}{80.77}}     \\
        \textbf{Average}  & 98.1/95.77/73.6/76.4/60.49                                                                                         & 98.63/\underline{\textcolor{blue}{96.81}}/77.59/80.9/\underline{\textcolor{blue}{65.99}} & 98.72/96.73/78.28/\underline{\textcolor{blue}{80.96}}/64.91                                                                                                         & \underline{\textcolor{blue}{98.88}}/96.74/\underline{\textcolor{blue}{78.74}}/80.88/65.32                                                                                          & \textbf{\textcolor{red}{99.11}}/\textbf{\textcolor{red}{97.73}}/\textbf{\textcolor{red}{80.28}}/\textbf{\textcolor{red}{82.75}}/\textbf{\textcolor{red}{68.42}}     \\
        \bottomrule
      \end{tabular}
    }
  \end{center}
\end{table*}

\subsubsection{Results on BTAD and KSDD2}
Similarly, we compare the proposed method with other generation algorithms in terms of the
AD performance on the other two well-known AD datasets. The accuracies obtained by DeSTseg
are summarized in Tab.~\ref{tab:btad} and Tab.~\ref{tab:ksdd2} for BTAD
\cite{mishra2021btad} and KSDD2 \cite{bovzivc2021mixed}, respectively. One can again
observe the remarkable superiority of the proposed method in the two tables. The artificial
defects simulated by AdaBLDM lead to the best performances for all the involved AD metrics, on
both two datasets. 

% BTAD DeSTseg
\begin{table*}[h!]
  \label{tab:btad}
  \begin{center}
    \caption{The Anomaly Detection and Localization Performance of DeSTseg on the
    BTAD dataset, Pixel-Auc/PRO/AP/IAP/IAP90, the best and the second-best numbers are
    shown in red and blue, respectively.}
    \resizebox{\linewidth}{!}{
      \begin{tabular}{c c c c c c c}
        \toprule
       \textbf{Category} & \makecell{\textbf{DESTSEG}\cite{zhang2023destseg}\NL (CVPR2023)}&\makecell{\textbf{DFM}\cite{AAAI2023DFM} \NL (AAAI2023)}& \makecell{\textbf{DCDGANc-StarGAN}\cite{WACVDCGAN} \NL (CVPRW2023)} &\makecell{\textbf{DCDGANc-StyleGAN}\cite{WACVDCGAN}\NL (CVPRW2023)}& \makecell{\textbf{\OUR} \NL (ours) }\\
        \hline
        01                & \underline{\textcolor{blue}{96.25}}/\underline{\textcolor{blue}{82.5}}/\textbf{\textcolor{red}{45.74}}/\textbf{\textcolor{red}{39.93}}/\underline{\textcolor{blue}{26.95}} & 95.97/74.17/37.13/31.71/14.92                                                                                           & 95.33/71.39/35.37/31.2/11.32                                & 95.64/82.34/38.15/36.24/26.78                                                            & \textbf{\textcolor{red}{96.84}}/\textbf{\textcolor{red}{84.25}}/\underline{\textcolor{blue}{43.22}}/\underline{\textcolor{blue}{39.04}}/\textbf{\textcolor{red}{27.93}} \\
        02                & 95.73/57.99/60.58/40.82/7.96                                                                                                                                               & \underline{\textcolor{blue}{96.15}}/\underline{\textcolor{blue}{64.78}}/62.73/47.46/\underline{\textcolor{blue}{12.61}} & 94.67/60.65/55.6/\textbf{\textcolor{red}{49.6}}/10.63       & 94.39/61.32/\underline{\textcolor{blue}{64.72}}/45.09/8.46                               & \textbf{\textcolor{red}{96.55}}/\textbf{\textcolor{red}{65.46}}/\textbf{\textcolor{red}{70.16}}/\underline{\textcolor{blue}{47.99}}/\textbf{\textcolor{red}{13.24}}     \\
        03                & \textbf{\textcolor{red}{99.47}}/\textbf{\textcolor{red}{98.01}}/\underline{\textcolor{blue}{44.29}}/35.87/17.54                                                            & 98.94/95.45/39.13/34.18/\textbf{\textcolor{red}{19.22}}                                                                 & 98.92/95.32/37.26/35.82/\underline{\textcolor{blue}{18.02}} & 99.27/97.43/43.5/\underline{\textcolor{blue}{36.58}}/16.2                                & \textbf{\textcolor{red}{99.47}}/\underline{\textcolor{blue}{97.97}}/\textbf{\textcolor{red}{45.91}}/\textbf{\textcolor{red}{36.59}}/17.95                               \\
        \textbf{Average}  & \underline{\textcolor{blue}{97.15}}/79.5/\underline{\textcolor{blue}{50.2}}/38.87/\underline{\textcolor{blue}{17.48}}                                                      & 97.02/78.13/46.33/37.78/15.58                                                                                           & 96.31/75.79/42.74/38.87/13.32                               & 96.43/\underline{\textcolor{blue}{80.36}}/48.79/\underline{\textcolor{blue}{39.3}}/17.15 & \textbf{\textcolor{red}{97.62}}/\textbf{\textcolor{red}{82.56}}/\textbf{\textcolor{red}{53.1}}/\textbf{\textcolor{red}{41.21}}/\textbf{\textcolor{red}{19.71}}          \\
        \bottomrule
      \end{tabular}
    }
  \end{center}
\end{table*}

% KSDD2 DeSTseg
\begin{table}[h!]
  \label{tab:ksdd2}
  \begin{center}
    \caption{The Anomaly Detection and Localization Performance of DeSTseg on the
    KSDD2 dataset, Pixel-Auc/PRO/AP/IAP/IAP90, the best and the second-best numbers are
    shown in red and blue, respectively.}
    \resizebox{\linewidth}{!}{
      \begin{tabular}{c c c c c c }
        \toprule
        \textbf{Method}                                                     & Pixel AUC                       & PRO                             & AP                              & IAP                             & IAP90                         \\
        \hline
        \makecell{\textbf{DESTSEG}\cite{zhang2023destseg}\NL (CVPR2023)}    & 89.45                           & 83.42                           & 52.3                            & 57.59                           & 1.29                          \\
        \makecell{\textbf{DFM}\cite{AAAI2023DFM} \NL (AAAI2023)}            & 86.69                           & 85.53                           & 47.38                           & 54.39                           & 1.26                          \\
        \makecell{\textbf{DCDGANc-StarGAN}\cite{WACVDCGAN} \NL (CVPRW2023)} & 93.08                           & 85.3                            & 46.53                           & 49.96                           & 1.69                          \\
        \makecell{\textbf{DCDGANc-StyleGAN}\cite{WACVDCGAN}\NL (CVPRW2023)} & 92.38                           & 84.83                           & 51.72                           & 55.65                           & 1.56                          \\
        % \textbf{\OUR} \textbf{(w.o Adapt.)} & \underline{\textcolor{blue}{96.01}} & \underline{\textcolor{blue}{91.87}} & \underline{\textcolor{blue}{68.19}} & \underline{\textcolor{blue}{73.34}} & \underline{\textcolor{blue}{4.91}} \\
        \makecell{\textbf{\OUR} \NL (ours) }                                & \textbf{\textcolor{red}{96.42}} & \textbf{\textcolor{red}{92.93}} & \textbf{\textcolor{red}{70.62}} & \textbf{\textcolor{red}{73.88}} & \textbf{\textcolor{red}{7.6}} \\
        \bottomrule
      \end{tabular}
    }
  \end{center}
\end{table}

\subsubsection{Image Quality}
Tab.~\ref{tab:kid_lpips} demonstrates the scores on the image quality of the comparing
generation algorithms. From the table, we can see that the proposed method achieves the
best KID performances for most subcategories (except Grid and Wood) as well as the overall
comparison. This is consistent with the AD performances reported in
Tab.~\ref{tab:destseg_mvtec} as the KID metric reflects the ``reality'' of the generated
images. 

\begin{table}[h!]
  \label{tab:kid_lpips}
  \begin{center}
    \caption{The KID$\times 10^{3}$@5k $\downarrow$ and LPIPS@5k $\uparrow$ of the generated images on MVTec
    AD, the best and the second-best numbers are shown in red and blue, respectively}
    \resizebox{1\linewidth}{!}{
    \begin{tabular}{c c c c c c}
      \toprule
      \textbf{Category} & \makecell{\textbf{DFM}\cite{AAAI2023DFM} \NL (AAAI2023)} & \makecell{\textbf{DCDGANc-StarGAN}\cite{WACVDCGAN} \NL (CVPRW2023)}    & \makecell{\textbf{DCDGANc-StyleGAN}\cite{WACVDCGAN}\NL (CVPRW2023)}    & \makecell{\textbf{\OUR} \NL (ours) }\\
      \hline
              \textbf{Hazelnut}  & 4.74/\textbf{\textcolor{red}{0.188}}      & \underline{\textcolor{blue}{1.76}}/0.16                                & 1.97/0.167                                                             & \textbf{\textcolor{red}{1.72}}/\underline{\textcolor{blue}{0.182}}  \\
        \textbf{Wood}      & 48.47/\textbf{\textcolor{red}{0.172}}     & \underline{\textcolor{blue}{7.91}}/0.046                               & 8.74/0.048                                                             & \textbf{\textcolor{red}{2.74}}/\underline{\textcolor{blue}{0.076}}  \\
        \textbf{Capsule}   & 3.66/0.018                                & \underline{\textcolor{blue}{1.87}}/\underline{\textcolor{blue}{0.026}} & \textbf{\textcolor{red}{1.08}}/\underline{\textcolor{blue}{0.026}}     & 4.14/\textbf{\textcolor{red}{0.027}}                                \\
        \textbf{Leather}   & 61.35/\textbf{\textcolor{red}{0.199}}     & \underline{\textcolor{blue}{9.23}}/\underline{\textcolor{blue}{0.088}} & 13.6/0.078                                                             & \textbf{\textcolor{red}{5.85}}/0.032                                \\
        \textbf{Grid}      & \underline{\textcolor{blue}{4.00}}/0.061  & \textbf{\textcolor{red}{2.04}}/\underline{\textcolor{blue}{0.073}}     & 7.21/\textbf{\textcolor{red}{0.087}}                                   & 5.38/0.07                                                           \\
        \textbf{Tile}      & \underline{\textcolor{blue}{16.74}}/0.015 & 23.16/0.036                                                            & 21.5/\textbf{\textcolor{red}{0.05}}                                    & \textbf{\textcolor{red}{12.83}}/\underline{\textcolor{blue}{0.046}} \\
        \textbf{Carpet}    & 3.64/0.082                                & 4.29/\textbf{\textcolor{red}{0.095}}                                   & \underline{\textcolor{blue}{3.61}}/\underline{\textcolor{blue}{0.088}} & \textbf{\textcolor{red}{3.55}}/0.073                                \\
        \textbf{Average}   & 20.37/\textbf{\textcolor{red}{0.105}}     & \underline{\textcolor{blue}{7.18}}/0.075                               & 8.25/\underline{\textcolor{blue}{0.078}}                               & \textbf{\textcolor{red}{5.17}}/0.072                                \\
      \bottomrule
    \end{tabular}
  }
      \end{center}
\end{table}

In contrast, it is interesting to see that our method lead to poor LPIPS scores compared with
other generation algorithms. Recall that either the editing process in the denoising stage
or the online decoder adaptation prevents the generated image from being too diverse
from the source image and the desired defect. \emph{In this way, we claim that the poor
image diversity might be a disadvantage of the proposed algorithm from the traditional
perspective while it also implies a better-controlled generation process that can benefit
the following training stage of AD models.} 

% Result
\subsection{Qualitative Results}
\begin{figure*}[ht]
  \centering{
    \includegraphics[scale=0.7]{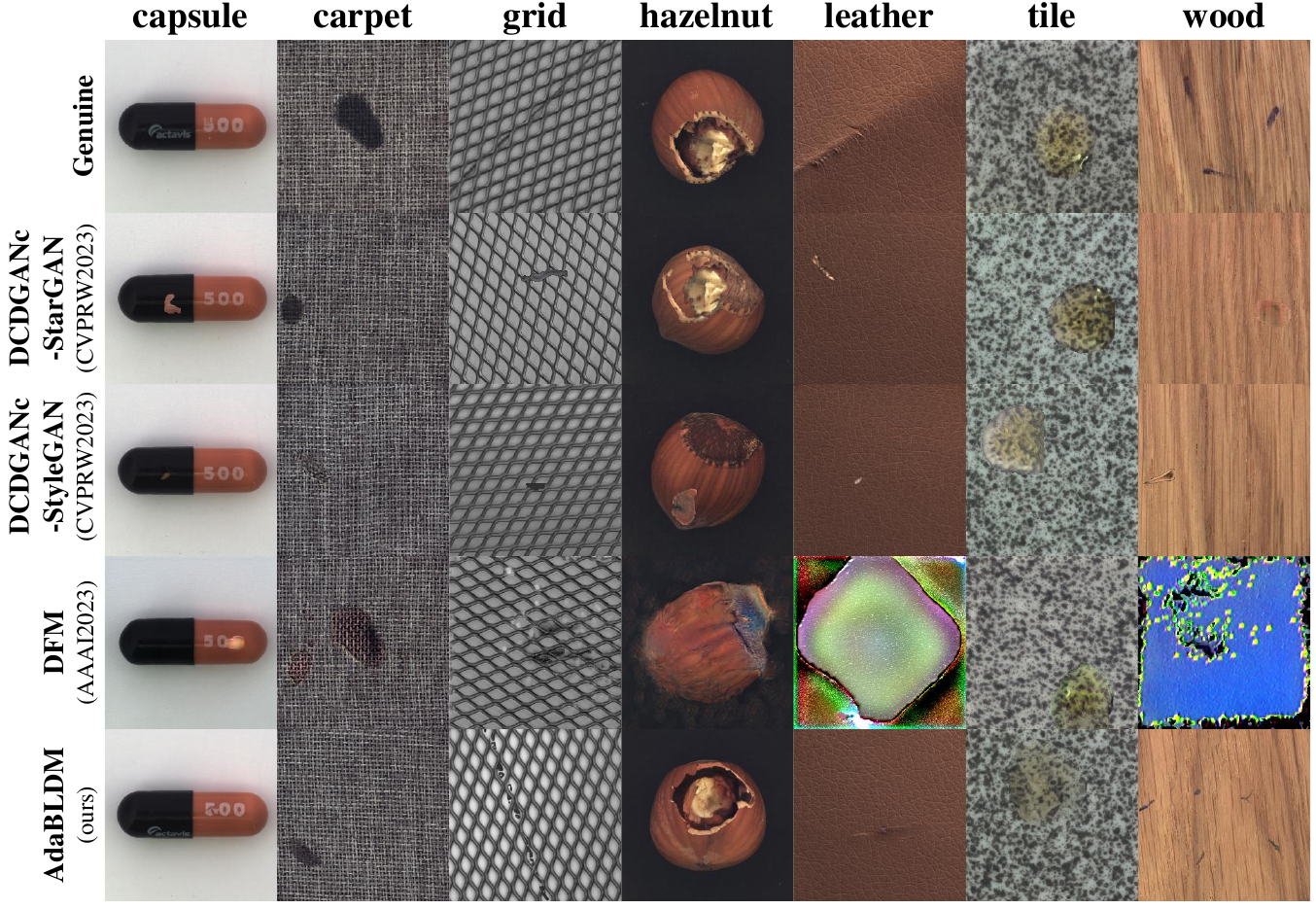}
    \caption{The genuine and synthetic defective samples on MVTec AD dataset. From top to bottom, the samples with genuine defects; the samples generated by using DCDGANc-StarGAN; the samples generated by using DCDGANc-StyleGAN; the samples generated with DFM and the samples generated by using our method.}
    \label{fig:gen_mvtec}
  }
\end{figure*}

To offer readers a more direct impression of the proposed image generation algorithm, We
hereby depict the generated defects by using the comparing methods in
Fig.~\ref{fig:gen_mvtec}. One can see the proposed method, either with or without the online
adaptation stage, can achieve significantly more lifelike images compared with other
methods. In particular, only our method can generate the ``crack'' defects on the hazelnut
subcategory. On the contrary, the DFM algorithm, though can generate synthetic defective
samples with good quality in some scenarios, could fail in generation for some other
cases. 

\begin{figure*}[ht]
  \centering{
    \includegraphics[scale=0.6]{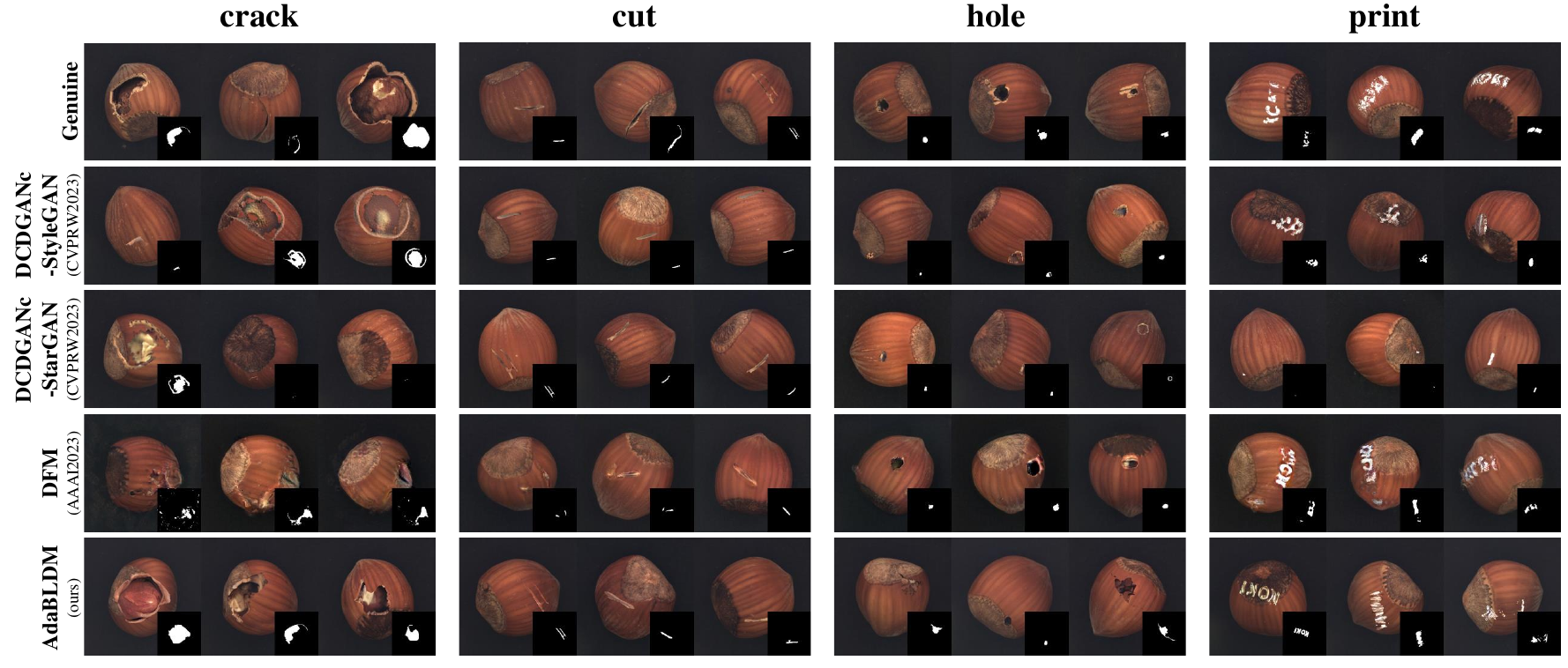}
    \caption{The fine-grained comparison in the generation quality on the Hazelnut subcategory of MVTec AD. Different kinds of defects are mimicked by the involved generation algorithms and their performances can be compared on a more detailed level. }
    \label{fig:gen_fine}
  }
\end{figure*}

\begin{figure*}[ht]
  \centering{
    \includegraphics[scale=0.5]{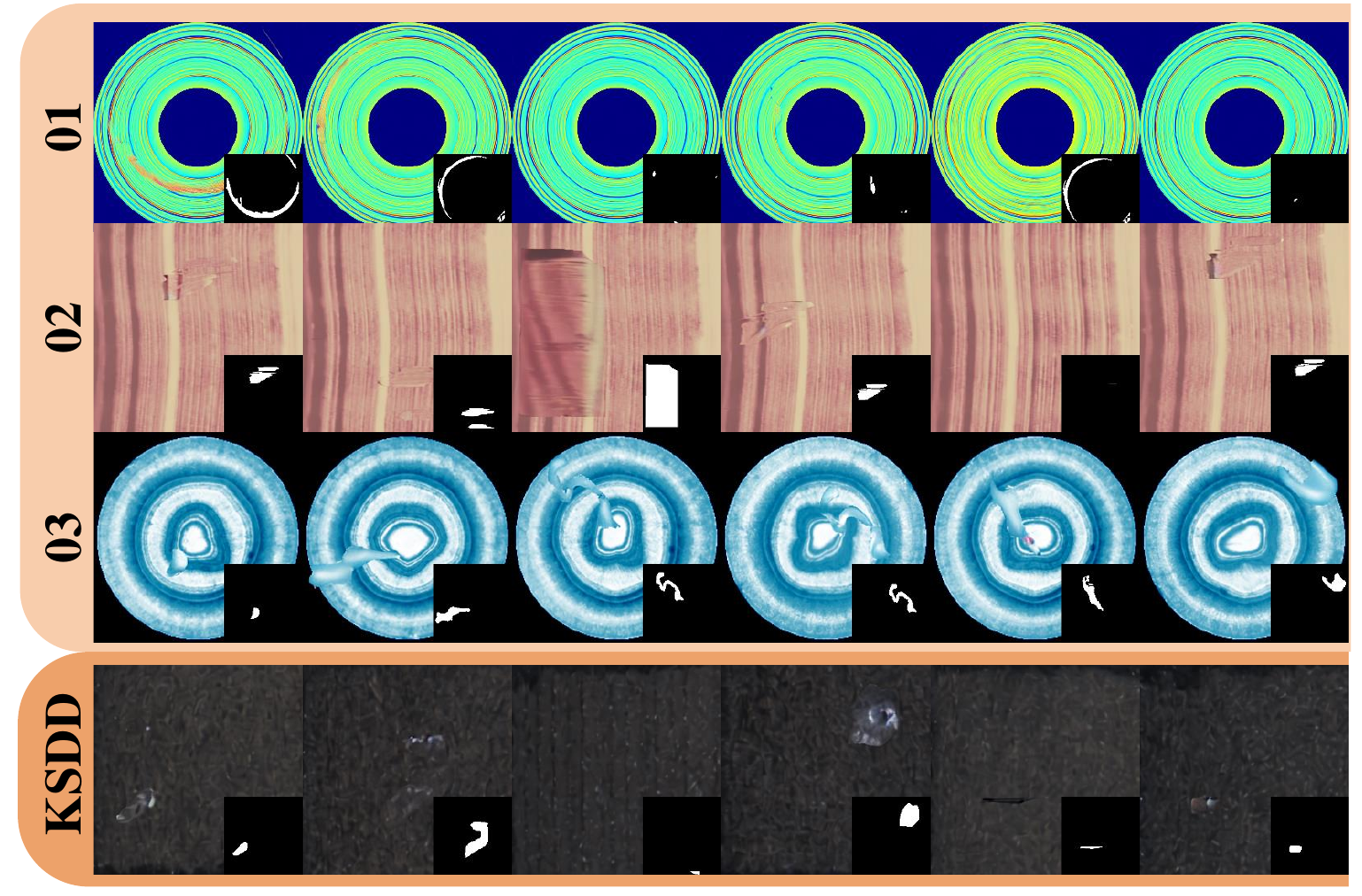}
    \caption{The generation outputs of AdaBLDM on BTAD and KSDDV2 datasests.}
    \label{fig:gen_two}
  }
\end{figure*}

We further compare with existing methods in fine-grained details, and experimental results are reported in
Fig.~\ref{fig:gen_fine}. As shown in Fig.~\ref{fig:gen_fine}, the proposed method is more well-controlled in generating the defect regions with the specific category and location. In addition, the generated images with synthetic defects on BTAD and KSDD2 are  shown
in Fig.~\ref{fig:gen_two}. It can be seen that besides MVTec AD, the image quality of AdaBLDM
is maintained for the two different datasets. 

\begin{figure*}[ht]
  \centering{
    \includegraphics[scale=0.8]{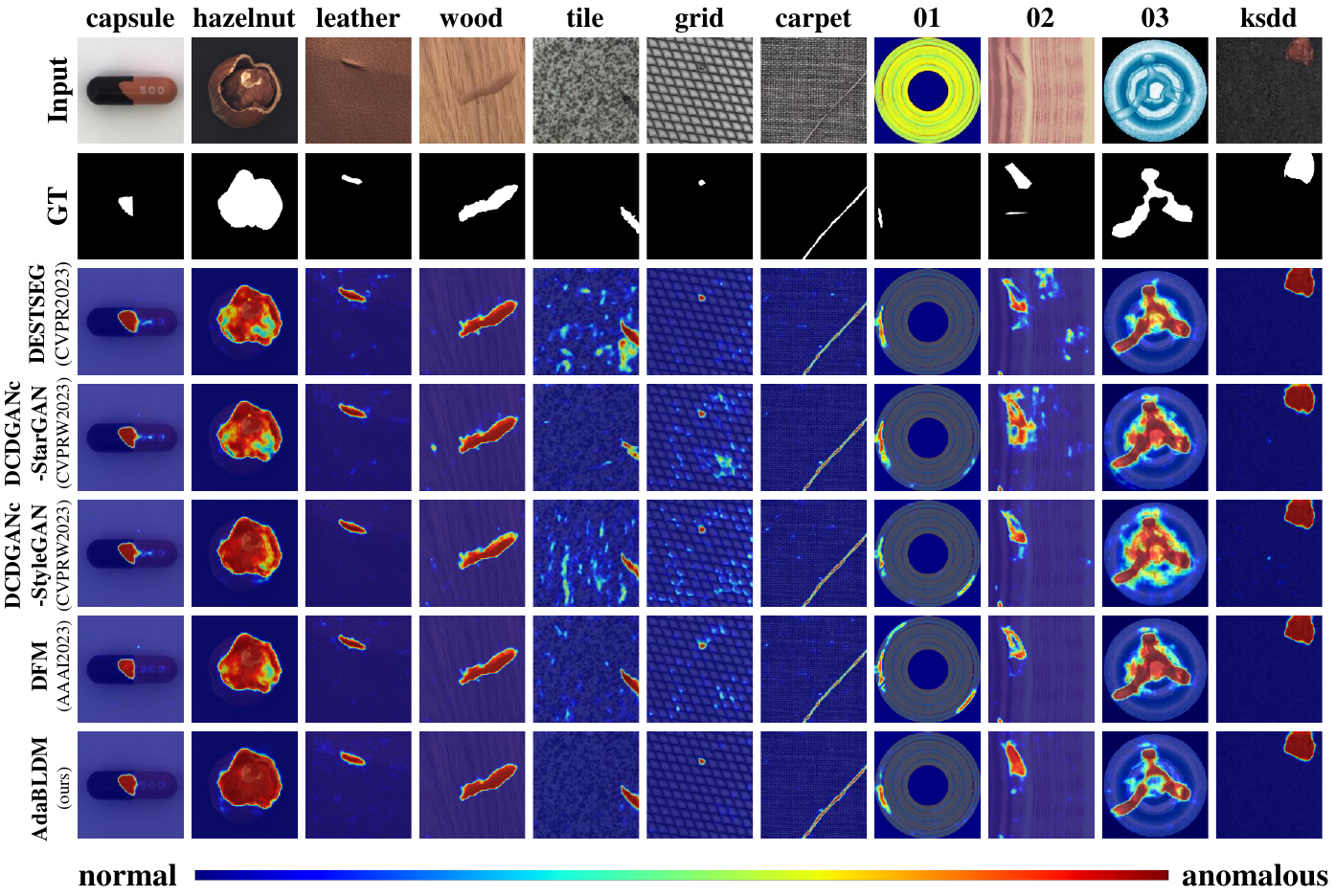}
    \caption{Test results of different DeSTseg models learned based on the training sets obtained by using different generation methods. All the involved subcategories in this work ($7$ from MVTec AD, $3$ from BTAD, and the only one in KSDDV2) are shown here.}
    \label{fig:destseg_mvtec}
  }
\end{figure*}

Finally, as a qualitative complement to Tab.~\ref{tab:destseg_mvtec},
Fig.~\ref{fig:destseg_mvtec} shows the heat map of anomaly predicted by the DeSTseg
algorithm. We can see that high-quality simulated synthetics benefits the
AdaBLDM-equipped algorithm which can output competitive heat maps for anomaly detection and localization. 

% Ablation
\subsection{Ablation Study}
In this section, we systematically evaluate the contribution of each component in our method. The following experiments are conducted:
\begin{itemize}
  \item \textbf{Pretrain with domain knowledge} \\
    As introduced in Sec.~\ref{subsec:training}, the denoising model
    ${\epsilon}_{\theta}(\cdot)$ is pretrained on the domain data. If this
    module is removed, AdaBLDM will rely on the prior knowledge from the LAION-5B
    dataset \cite{schuhmann2022laion}, which might not highly correlate with the
    domain of industrial AD. Note that the pretraining is strictly independent to the
    current subcategory to maintain the fair comparison. 
  \item \textbf{Image Editing} \\
    Different from the conventional blended latent diffusion model, we propose to perform
    the intent editing in both latent space and pixel space (see Alg.~\ref{alg:editing}).
    If this module is removed, the editing is only performed in the latent space.  
  \item \textbf{Online Decoder Adaptation} \\
    The online decoder adaptation is introduced in Alg.~\ref{alg:online}. The absence of
    this module results in a fixed image decoder. 
\end{itemize}
The ablation results are summarized in Tab.~\ref{table:ablation}. It is clear that as each
module mentioned above is added to the algorithm, the corresponding AD performance
increases steadily. In particular, the total performance improvements are $9.78\%$ in AP,
$14.46\%$ in IAP, and $13.98\%$ in IAP90, respectively.

\begin{table*}[h!]
  \begin{center}
    \caption{Ablation Study of AdaBLDM}
    \label{table:ablation}
    \resizebox{.7\linewidth}{!}{
      \begin{tabular}{c c c c c c c c  }
        \toprule
        \multicolumn{3}{c}{\textbf{Modules}} & \multicolumn{5}{c}{\textbf{Performance}}                                                                                                                                                                                                            \\

        \cmidrule(lr){1-3}                           \cmidrule(lr){4-8}
        Pretrain & Image Editing & Adaptation & Pixel auc                           & Pro                                 & AP                                  & IAP                                 & IAP90                               \\
        \ding{55}                            & \ding{55}                                & \ding{55}  & 96.51                               & 92.18                               & 70.50                               & 68.29                               & 54.44                               \\
        \checkmark                           & \ding{55}                                & \ding{55}  & 97.23                               & 93.14                               & 71.21                               & 69.30                               & 58.32                               \\
        \checkmark                           & \checkmark                               & \ding{55}  & \textbf{\textcolor{red}{99.12}}     & \underline{\textcolor{blue}{97.49}} & \underline{\textcolor{blue}{80.25}} & \underline{\textcolor{blue}{82.36}} & \underline{\textcolor{blue}{67.65}} \\
        \checkmark                           & \checkmark                               & \checkmark & \underline{\textcolor{blue}{99.11}} & \textbf{\textcolor{red}{97.73}}     & \textbf{\textcolor{red}{80.28}}     & \textbf{\textcolor{red}{82.75}}     & \textbf{\textcolor{red}{68.42}}     \\
        \bottomrule
      \end{tabular}
    }
  \end{center}
\end{table*}

% \begin{figure*}[ht]
%   \centering{
%     \includegraphics[scale=0.6]{figs/destseg_result_categories_details}
%     \caption{all categories performents details on MVTec AD by using DeSTseg.}
%     \label{fig:destseg_mvtec_details}
%   }
% \end{figure*}

\section{Conclusion \& future work}
\label{sec:conclusion}
In this paper, we propose an advanced method to generate realistic defective samples for industrial anomaly detection and localization. Inspired by the diffusion model, we customize the Blended Latent Diffusion Model (BLDM) to be well-suited to this specific task. The proposed algorithm, termed AdaBLDM, can generate better synthetic
samples compared with existing methods based on conventional ``cut-and-paste'' schemes or Generative Adversarial Networks (GANs). By applying the AdaBLDM in anomaly detection data
augmentation, we achieve a state-of-the-art performance of AD, thanks to the well-controlled defective regions aligned at the pixel level.
% Also, the proposed algorithm can be directly used as an effective data
% augmentation method in real-world applications for defect generalization tasks. 
In the future, we explore high-quality synthetic defective samples by introducing finer controlling signals such as the language
prompts describing the details (shape, size, etc.) of the demanding defects. Another
appealing direction of future work is to speed up the denoising process of the diffusion
model so that more defective samples can be generated given a limited time
budget, which is a crucial constraint in most practical scenarios.

\bibliography{anomaly_detection}
\bibliographystyle{IEEEtran}

\end{document}